\documentclass[conference]{IEEEtran}
\IEEEoverridecommandlockouts

\usepackage{geometry}
\usepackage{verbatim} 
\usepackage{caption}
\usepackage{graphics} 
\usepackage{graphicx}
\usepackage{times} 
\usepackage{amssymb}  
\usepackage{amsmath}
\usepackage{xcolor}
\usepackage{hyperref}
\usepackage{booktabs}
\usepackage{bbm}
\usepackage{array} 
\usepackage{float}  
\usepackage{balance} 
\usepackage[symbol]{footmisc}
\usepackage[numbers,sort&compress,square,comma]{natbib}

\newcommand{\figref}[1]{Fig.~\ref{fig:#1}}
\newcommand{\tableref}[1]{Table ~\ref{tab:#1}}

\title{\vspace{0.25in}Enhancing Metabolic Syndrome Prediction with Hybrid Data Balancing and Counterfactuals}

\author{Sanyam Paresh Shah$^1$, Abdullah Mamun$^{1,2,*}$, Shovito Barua Soumma$^1$, Hassan Ghasemzadeh$^1$

\thanks{$^1$College of Health Solutions, Arizona State University, Phoenix, AZ 85054.} \\
\thanks{$^2$School of Computing and Augmented Intelligence, Arizona State University, Tempe, AZ 85281.} \\
\thanks{$^*$Corresponding author.}
\thanks{\{sshah174, a.mamun, shovito, hassan.ghasemzadeh\}@asu.edu%
}
}

\begin{document}

\maketitle

\begin{abstract}
Metabolic Syndrome (MetS) is a cluster of interrelated risk factors that significantly increases the risk of cardiovascular diseases and type 2 diabetes. Despite its global prevalence, accurate prediction of MetS remains challenging due to issues such as class imbalance, data scarcity, and methodological inconsistencies in existing studies. In this paper, we address these challenges by systematically evaluating and optimizing machine learning (ML) models for MetS prediction, leveraging advanced data balancing techniques and counterfactual analysis. Multiple ML models, including XGBoost, Random Forest, TabNet, etc., were trained and compared under various data balancing techniques such as random oversampling (ROS), SMOTE, ADASYN, and CTGAN. Additionally, we introduce \textit{MetaBoost}\footnote[3]{The source code is available at: \href{https://github.com/asusanyamshah/MetaBoost}{\textcolor{blue}{github.com/asusanyamshah/MetaBoost}}}, a novel hybrid framework that integrates SMOTE, ADASYN, and CTGAN, optimizing synthetic data generation through weighted averaging and iterative weight tuning to enhance the model's performance (achieving up to a 1.87\% accuracy improvement over individual balancing techniques). A comprehensive counterfactual analysis is conducted to quantify the feature-level changes required to shift individuals from high-risk to low-risk categories. The results indicate that blood glucose (50.3\%) and triglycerides (46.7\%) were the most frequently modified features, highlighting their clinical significance in MetS risk reduction. Additionally, probabilistic analysis shows elevated blood glucose (85.5\% likelihood) and triglycerides (74.9\% posterior probability) as the strongest predictors.  This study not only advances the methodological rigor of MetS prediction but also provides actionable insights for clinicians and researchers, highlighting the potential of ML in mitigating the public health burden of metabolic syndrome.
\end{abstract}


\section{Introduction}
Metabolic Syndrome (MetS), characterized by a cluster of risk factors including abdominal obesity, dyslipidemia, hypertension, and insulin resistance, significantly elevates the risk of cardiovascular diseases (CVD) and type 2 diabetes mellitus (T2DM)~\cite{ford2002prevalence, alberti2009harmonizing}. With a global prevalence exceeding 25\% in adults, MetS poses a substantial public health burden, underscoring the need for early and accurate risk prediction to enable timely interventions~\cite{saklayen2018global}. Traditional diagnostic frameworks, such as the National Cholesterol Education Program’s Adult Treatment Panel III (NCEP ATP III), rely on threshold-based criteria for individual risk factors, which may fail to capture nuanced interactions between variables or adapt to population-specific variations~\cite{grundy2005diagnosis}.

Challenges in healthcare datasets, such as missing or imbalanced data and small sample sizes, require the development of robust predictive models for accurate inference~\cite{ mamun2022designing, soummassl,fogsense,mamun2025glucolens, hussein2023energy}. To address these challenges, techniques such as random oversampling (ROS), Synthetic Minority Oversampling Technique (SMOTE)~\cite{chawla2002smote}, Adaptive Synthetic Sampling (ADASYN) \cite{he2008adasyn}, and generative models like Conditional Tabular Generative Adversarial Networks (CTGAN) \cite{xu2019modeling}, BIDC2 \cite{zhai2022binary}, and AIMEN \cite{mamun2024use} have been employed for data balancing. During synthetic data generation, it is important not to add too much post-generation filitering as it can cause the distribution of the training set to be different than that of the test set \cite{mamun2024use}. However, the comparative efficacy of these methods, particularly in combination, remains underexplored in the context of MetS prediction. 

Recent advances in ML offer the transformative potential for improving MetS prediction by integrating heterogeneous clinical, demographic, and laboratory data into holistic risk assessments~\cite{chen2019machine,10054906}.  Existing works on hyperglycemia and hypoglycemia detection demonstrate the potential of leveraging machine learning and simulation-based approaches for personalized predictions and tailored interventions in metabolic health~\cite{arefeen2023glysim, shroff2023glucoseassist, arefeen2024glyman}. Existing studies often face limitations such as class imbalance, data scarcity, and inconsistent preprocessing protocols, which can lead to inflated performance metrics and reduced generalizability~\cite{Soumma_Mamun_Ghasemzadeh_2025,johnson2019survey,azghan2025cudle}. While ML models like XGBoost, Random Forest, and deep learning architectures have demonstrated promise in metabolic health analytics, rigorous comparative analyses under standardized preprocessing and evaluation frameworks are lacking.
\begin{figure*}[t]
    \centering
    \includegraphics[width=0.99\linewidth]{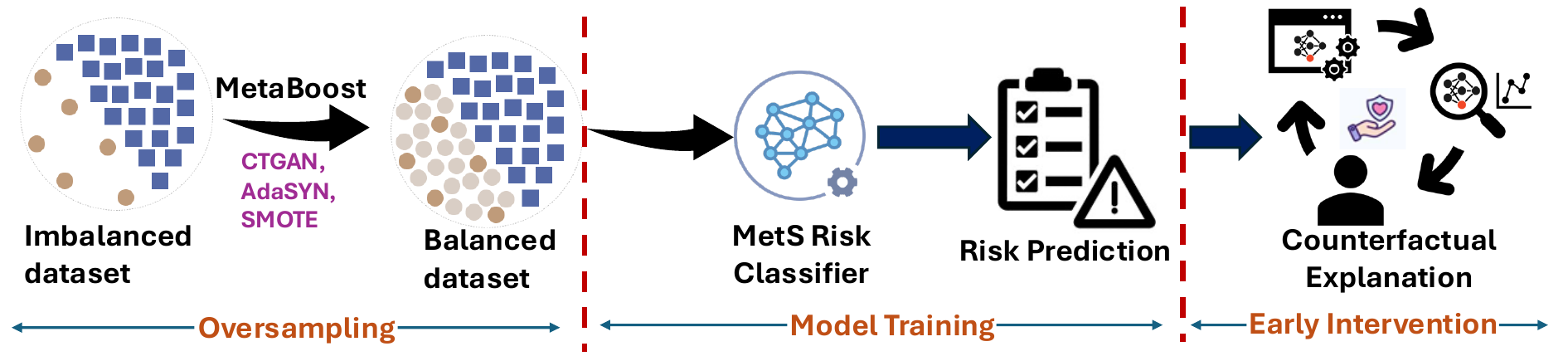}
    \caption{The MetaBoost framework integrates advanced data augmentation techniques (CTGAN, ADASYN, SMOTE) to address class imbalance in metabolic syndrome prediction. The pipeline consists of oversampling, model training, risk prediction, and counterfactual analysis, enabling early intervention for at-risk individuals.}
    \label{fig:framework}
    \vspace{-2mm}
\end{figure*}
To bridge these gaps, we investigate the following research question: \textit{How can hybrid machine learning approaches, leveraging advanced data balancing techniques and counterfactual analysis, enhance the early detection, risk stratification, and clinical interpretability of MetS prediction models?} Motivated by these gaps, this study aims to systematically evaluate and optimize ML models for MetS prediction, leveraging a comprehensive preprocessing pipeline and advanced data balancing techniques. Specifically, we address class imbalance through a combination of different methods for oversampling, such as SMOTE, ADASYN, and CTGAN, exploring both individual and hybrid approaches to synthetic data generation. To this end, we introduce \textit{\textbf{MetaBoost}}, a novel hybrid framework that integrates these techniques through weighted averaging and iterative weight tuning to optimize synthetic data generation and improve model robustness. Furthermore, we conduct a counterfactual analysis to elucidate the feature-level changes required to shift individuals from high-risk to low-risk categories, providing actionable insights for clinical interventions.

This work contributes: i) an evaluation of ML models (e.g., XGBoost, Random Forest, TabNet) under data balancing strategies like SMOTE and CTGAN; ii) a novel hybrid data balancing framework combining SMOTE, ADASYN, and CTGAN with optimized weighting; iii) counterfactual analysis to quantify feature changes and decision boundaries; and iv) a probabilistic risk analysis based on clinical thresholds for risk stratification.

By addressing these objectives, this study not only advances the methodological rigor of MetS prediction but also provides actionable insights for clinicians and researchers aiming to mitigate the public health burden of metabolic syndrome. Our findings highlight the potential of hybrid data balancing techniques and counterfactual analysis in enhancing the interpretability and generalizability of ML models in healthcare.

\section{Materials and Methods}

\subsection{Dataset}
The dataset used for this experiment comprises demographic, clinical, and laboratory measurements for 2,401 individuals, with 15 columns that entails 13 features, a sequence ID and the response variable indicating the presence or absence of metabolic syndrome. The 13 features include age, sex, marital status, income, race, waist circumference, BMI (body mass index), albuminuria (presence of albumin, a blood protein, in the urine), urine albumin–creatinine ratio, uric acid, blood glucose, HDL (high-density lipoprotein), and triglycerides.  This is a dataset derived from the National Health and
Nutrition Examination Survey (NHANES) database. We refer the readers to the original data collection paper for further details about the dataset and the data collection procedure \cite{trigka2023predicting}. We conducted a risk analysis to assess the prevalence of metabolic syndrome in the dataset and to compute likelihood and posterior probabilities for specific feature combinations.

\subsection{Data Preprocessing}
The preprocessing pipeline involved several critical steps to ensure the dataset was suitable for analysis. As a considerable number of rows (8.66\%) had missing values for marital status, this feature was removed from the input feature set. Categorical variables, such as ``sex'' and ``race'' were encoded numerically. For ``sex'', Male was encoded as 0 and Female as 1. For ``race'', the categories were mapped as follows: White (0), Asian (1), Black (2), Mexican American (3), Hispanic (4), and Other (5).

Missing values in the columns ``Income'', ``WaistCirc'', and ``BMI'' were imputed using mean imputation. The dataset was then split into training and testing sets, with the testing set comprising 33\% of the total data. It was ensured that the test set contained an equal number of instances for both classes of the response variable (presence or absence of metabolic syndrome) to avoid misinterpretation of the evaluation results.

\subsection{Model Selection and Evaluation}
Multiple machine learning models were employed for the analysis, including Random Forest Classifier, Decision Tree Classifier, XGBoost Classifier ~\cite{chen2016xgboost}, Logistic Regression, Multilayer Perceptron (MLP) Classifier, and TabNet Classifier ~\cite{arik2021tabnet}. Model performance was evaluated using standard metrics: accuracy, precision, recall, and F1 score. These metrics were chosen to provide a comprehensive assessment of model performance, particularly in the context of class imbalance. The metrics were recorded by testing the model 3 times in each case, and recording the average performance. 

\subsection{Analysis of Oversampling Strategies}
Two distinct strategies were employed to address class imbalance. The first approach did not use any oversampling, and models were trained on the original imbalanced dataset. In the second approach, random oversampling was applied only to the training set, while the testing set remained unchanged. For each strategy, the dataset was split such that the testing set constituted 33\% of the total data, with equal representation of both classes in the test set. Performance metrics (accuracy, precision, recall, and F1 score) were recorded for each model and strategy combination.

\subsection{Analysis of Class Imbalance Handling Methods}
To further mitigate class imbalance, advanced techniques such as Synthetic Minority Oversampling Technique (SMOTE), Adaptive Synthetic Sampling (ADASYN), and Conditional Tabular Generative Adversarial Networks (CTGAN) were employed ~\cite{douzas2018improving}. These methods were tested individually and in hybrid combinations. For hybrid approaches, synthetic data points were generated by calculating a weighted average of the nearest pairs from the training dataset. For combinations involving two methods, the weights for each method were systematically varied in increments of 0.05, with 20 iterations performed for each combination. For hybrid approaches combining two methods that involved CTGAN, CTGAN was trained for 300 epochs. 

For the combination of SMOTE, ADASYN, and CTGAN, the weights were adjusted iteratively. Initially, the weight of SMOTE was set to 0, while CTGAN's weight was incremented by 0.05 in each iteration. The weight of ADASYN was calculated as the remainder to ensure the weights summed to 1. This process continued until the weight of SMOTE reached 1, resulting in 235 unique combinations. Due to computational constraints, CTGAN was trained for 100 epochs in this scenario.

\subsection{Counterfactual Analysis} 
Counterfactual examples were generated using the Nearest Instance Counterfactual Explanations (NICE) \cite{brughmans2024nice} algorithm to identify the minimal feature modifications required to shift individuals between the positive class (MetS present) and negative class (MetS absent). The counterfactual search was formulated as an optimization problem:
\begin{equation}
    \vspace{-1mm}
    \arg\min_{x'} \mathcal{D}(x, x') + \lambda \cdot \mathcal{C}(x')
\end{equation}

where $x$ represents the original instance, and $x'$ is the counterfactual example. 
$\mathcal{D}(x, x')$ measures the feature-wise distance between instances, computed using the L1 norm. 
$\mathcal{C}(x')$ ensures that $x'$ belongs to the opposite class of $x$ (i.e., $f(x) \neq f(x')$).
$\lambda$ is a regularization parameter balancing proximity and classification change.

The normalized average distance, standard deviation, average feature changes, and percentage of altered features were computed. To visualize decision boundaries, a Random Forest Classifier was applied to PCA-transformed, standardized data. A mesh grid covered the PCA data range, and predictions were plotted to highlight the separation between original and counterfactual instances.

\section{Results}
\subsection{Probability Analysis}

\subsubsection{Prior Probability}
Prior probability represents the baseline understanding of the prevalence of metabolic syndrome in the dataset before considering any specific risk factors. It is calculated by dividing the total number of metabolic syndrome cases by the total number of data points in the dataset. In this analysis, the prior probability was 0.342, indicating that 34.2\% of the entries had metabolic syndrome. This prevalence aligns with findings from Ford et al. (2002), who reported similar rates in US adults, particularly in certain age groups and demographic categories~\cite{ford2002prevalence} 
\begin{table}[h]
\centering
\caption{Likelihood of various symptoms in the presence of metabolic syndrome.}
\begin{tabular}{lll}
\toprule
Parameter  & Likelihood \% \\
\midrule
glucose\_risk  & 85.5\% \\
bmi\_risk  & 62.4\% \\
triglycerides\_risk  & 54.7\% \\
uralbcr\_risk  & 20.3\% \\
albuminuria\_risk  & 15.9\% \\
\bottomrule
\label{tab:likelihood}
\end{tabular}
\end{table}

\begin{table}[h]
\centering
\caption{Posterior probabilities of metabolic syndrome in the presence of various risk factors.}
\begin{tabular}{ll}
\toprule
Parameter  & Posterior \% \\
\midrule
triglycerides\_risk  & 74.9\% \\
bmi\_risk  & 59.9\% \\
glucose\_risk  & 58.7\% \\
uralbcr\_risk  & 53.5\% \\
albuminuria\_risk & 51.6\% \\
\bottomrule
\label{tab:posterior}
\end{tabular}
\end{table}

\subsubsection{Likelihood Analysis}
Likelihood calculations provide crucial insights into the relationship between specific risk factors and metabolic syndrome. The likelihood represents the probability of observing a particular risk factor in individuals who have metabolic syndrome. For example, as shown in \tableref{likelihood}, this analysis revealed that elevated blood glucose ($\geq$100 mg/dL) had the highest likelihood at 85.5\%, meaning that among individuals with metabolic syndrome, 85.5\% exhibited high blood glucose levels. This threshold was chosen based on the American Diabetes Association's 2021 guidelines, which identify fasting glucose $\geq$100 mg/dL as indicative of prediabetes and increased metabolic risk~\cite{ada2021}. Similarly, obesity (BMI $\geq$30 kg/m²) showed a likelihood of 62.4\%, supporting the WHO Technical Report Series 894 (2000) findings on the strong association between obesity and metabolic syndrome~\cite{who2000}. The International Diabetes Federation's 2006 consensus provided the evidence for selection of our waist circumference thresholds ($\geq$94 cm for men, $\geq$80 cm for women), while the NCEP ATP III guidelines informed our HDL cholesterol ($<$40 mg/dL for men, $<$50 mg/dL for women) and triglyceride ($\geq$150 mg/dL) thresholds, with triglycerides showing a likelihood of 54.7\%~\cite{idf2006, ncep2002}.  

Age-specific thresholds were particularly important in this analysis. Following Ford et al.'s (2002) findings, the male age threshold was set at 40 years, as their research demonstrated a significant increase in metabolic syndrome risk at this age~\cite{ford2002prevalence}. For women, the age of 51 years was chosen based on Carr's (2003) research linking metabolic syndrome risk to menopausal transition, which typically occurs around this age~\cite{carr2003}. Kidney function markers were included based on KDIGO 2012 guidelines, with urinary albumin-to-creatinine ratio $\geq$30 mg/g indicating kidney damage, showing likelihoods of 20.3\% for elevated UrAlbCr and 15.9\% for albuminuria~\cite{kdigo2012}.

\begin{table*}[]
\centering
\caption{Performance of models across different random oversampling (ROS) strategies. Values in the first section show performance without any ROS. The best values are highlighted in bold. Acc: Accuracy, Pre: Precision, Rec: Recall, F1: F1 Score. RF: Random Forest, DT: Decision Tree, LR: Logistic Regression, TNet: TabNet.}
\label{tab:my_label}

\scalebox{1}{\begin{tabular}{llllllll}
\toprule
     && RF    & DT    & XGB   & LR    & MLP   & TNet  \\
\midrule
     & Acc & 0.804 & 0.815 & \textbf{0.843} & 0.744 & 0.545 & 0.673 \\
     & Pre & 0.931 & 0.879 & \textbf{0.936} & 0.920 & 0.914 & 0.675 \\
Without ROS & Rec & 0.656 & 0.732 & 0.736 & 0.534 & 0.270 & \textbf{0.920} \\
     & F1 & 0.770 & 0.799 & \textbf{0.824} & 0.676 & 0.417 & 0.779 \\
\midrule
     & Acc & 0.827 & 0.815 & \textbf{0.859} & 0.798 & 0.629 & 0.811 \\
ROS on & Pre & \textbf{0.917} & 0.871 & 0.913 & 0.775 & 0.579 & 0.769 \\
Training Set & Rec & 0.719 & 0.741 & 0.793 & 0.838 & \textbf{0.945} & 0.890 \\
     & F1 & 0.806 & 0.801 & \textbf{0.849} & 0.805 & 0.718 & 0.825 \\
\bottomrule
\end{tabular}}
\label{tab:accuracies_diff_strategies}
\vspace{-2.5mm}
\end{table*}

\subsubsection{Posterior Probabilities}
Posterior probabilities represent the refined probability of metabolic syndrome given the presence of specific risk factors, calculated using Bayes' Theorem. This calculation combines the prior knowledge (prior probability) with the likelihood to provide updated probabilities. The formula used is: (likelihood $\times$ prior probability) / probability of the feature in the entire population. As seen in \tableref{posterior}, analysis revealed that elevated triglycerides were the strongest predictor, with a posterior probability of 74.9\%, meaning that individuals with triglycerides $\geq$150 mg/dL had a 74.9\% probability of having metabolic syndrome. This finding aligns with NCEP ATP III's emphasis on triglycerides as a key metabolic syndrome component~\cite{ncep2002}.

Obesity (BMI) showed the second-highest posterior probability at 59.9\%, supporting WHO's identification of BMI $\geq$30 kg/m² as a significant risk factor. Elevated blood glucose, despite having the highest likelihood, ranked third in posterior probabilities at 58.7\%, suggesting that while common in metabolic syndrome patients, it may also occur frequently in the general population. Kidney function markers (UrAlbCr and albuminuria) showed posterior probabilities above 50\% (53.5\% and 51.6\% respectively), supporting Karalliedde and Viberti's (2005) findings on the relationship between kidney function and metabolic syndrome~\cite{karalliedde2005}.

The difference between likelihood and posterior probabilities highlights the importance of considering both the presence of risk factors in metabolic syndrome patients and their occurrence in the general population. For instance, while blood glucose showed the highest likelihood, its lower posterior probability suggests it may be less specific as a diagnostic indicator when considered independently. This comprehensive probabilistic approach, grounded in evidence-based thresholds from authoritative sources, provides valuable insights for both clinical practice and research in metabolic syndrome risk assessment.

\subsection{Accuracy Comparison}
The results shown in \tableref{accuracies_diff_strategies} are consistent with our hypothesis that training an imbalanced dataset, where no ROS or other balancing methods were performed, would yield lower accuracy and F1 score. The best test accuracy when ROS is not applied (0.843) is lower than the test accuracy when ROS is applied to the training set (0.859).

The results demonstrated noticeable variations in model performance across different random over-sampling strategies. One key observation is that the XGBoost classifier consistently showed the highest accuracy. However, MLP consistently performed the worst in terms of accuracy. This might be due to the complexity of MLP, which might require more data for MLP to perform well. MLP is also prone to overfitting, which might explain its poor metrics~\cite{rynkiewicz2012general}.

\subsection{Precision, Recall, and F1}
The XGBoost Classifier consistently achieves the highest precision and F1 score. In contrast, the MLP shows variable precision due to its sensitivity to input feature scaling and tendency to overfit~\cite{rynkiewicz2012general}. Notably, TabNet achieves the highest recall without ROS, thanks to its sequential attention mechanism, effective handling of imbalanced datasets, and hybrid approach combining decision trees and neural networks. These features enhance TabNet's ability to identify true positives and generalize well, while regularization techniques prevent overfitting. Conversely, the MLP exhibits the lowest F1 scores, particularly without ROS, due to its overfitting tendencies and the need for extensive tuning and large datasets~\cite{rynkiewicz2012general}.

\begin{table*}[htbp]
\centering
\caption{Comparison of test accuracy, precision, recall, and F1 score of hybrid data balancing approaches with their optimal weights. Data generation was applied to the training dataset, and the testing was done only on real data.}
\begin{tabular}{llllll}
\toprule
\textbf{Method}  & \textbf{Weights} & \textbf{Accuracy} & \textbf{Precision} & \textbf{Recall} & \textbf{F1} \\
\midrule
SMOTE & - & 0.868 & 0.889 & 0.840 & 0.864 \\
ADASYN & - & 0.855 & 0.872 & 0.833 & 0.852 \\
CTGAN & - & 0.866 & \textbf{0.913} & 0.810 & 0.858 \\
ADASYN+CTGAN  & (0.4, 0.6) & \textbf{0.871}  & 0.890  & \textbf{0.848}  & \textbf{0.868}  \\ 
SMOTE+CTGAN   & (0.5, 0.5) & 0.869  & 0.891  & 0.840  & 0.865  \\
SMOTE+ADASYN  & (0.75, 0.25) & 0.861  & 0.877  & 0.840  & 0.858  \\
SMOTE+CTGAN+ADASYN & (0.05,0.55,0.4) & 0.869  & 0.889  & 0.843  & 0.865  \\

\bottomrule
\end{tabular}
\label{tab:combined_results}
\vspace{-2mm}
\end{table*}

\subsection{Analysis with ROS on the Training Set}
The first approach of \tableref{accuracies_diff_strategies}, where ROS is not performed, keeps the training data imbalanced, and achieving a good performance on the test set is often impractical in this way. The other approach was implemented by first manually splitting the training and testing data so that the testing set contains 33\% of the data points from the original dataset. Both classes would be equally represented in the testing set. Then, random oversampling was performed on only the training data. Considering the data from this analysis, XGBoost has the highest accuracy (0.859) and F1 score (0.849). MLP had the lowest accuracy, precision, and F1 score but the highest recall. On the other hand, Random Forest had the highest precision.

\subsection{MetaBoost - Combining Multiple Data Balancing Methods}
When multiple balancing methods were combined using the methods described above, the highest F1 score was achieved by a combination of ADASYN and CTGAN with weights of \(0.4\) and \(0.6\), respectively. This combination yielded an accuracy of \(87.1\%\) and an F1 score of 0.868. The higher weight assigned to CTGAN (\(0.6\)) compared to ADASYN (\(0.4\)) indicates that CTGAN has contributed more than ADASYN to the generated synthetic training data in this context. Moreover, the hybrid data balancing method (F1 score: 0.868) outperforms the best result found with individual balancing methods (F1 score: 0.864).

For the combination of SMOTE and CTGAN, the highest F1 score (0.865) was achieved when both methods were assigned equal weights (\(0.5, 0.5\)). This suggests that CTGAN and SMOTE had equal contributions or priorities in the synthetic training data samples. Finally, with the training dataset combining SMOTE and ADASYN, SMOTE was assigned a higher weight of \(0.75\) compared to ADASYN's \(0.25\) to reach the highest F1 score (0.858) of that combination.

\begin{figure}[t]
    \centering
    \includegraphics[width=0.99\linewidth]{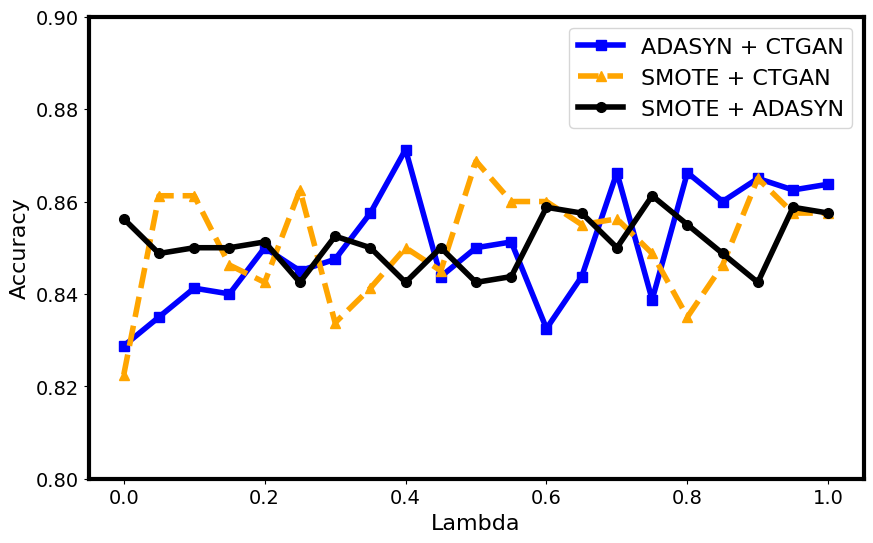}
    \caption{Comparison of accuracies of hybrid data balancing approaches across different weights. Weights: ($\lambda,1-\lambda$).}
    \label{fig:weights}
    \vspace{-2mm}
\end{figure}

The combination of all three methods (SMOTE, CTGAN, and ADASYN) with weights \(0.05\), \(0.55\), and \(0.4\), respectively, resulted in an accuracy of \(86.88\%\). This indicates that CTGAN still plays a dominant role in the combined approach, while SMOTE and ADASYN contribute to it to a lesser extent. Interestingly, here ADASYN is assigned a much higher weight than SMOTE. This might be because ADASYN might be providing synthetic samples that complement those generated by CTGAN and SMOTE effectively for this combination of weights. This complementary effect can enhance the overall diversity and quality of the synthetic data, leading to better model performance. There might also be interaction effects between the synthetic samples generated by the different methods that might explain the unpredictable nature of the weights that result in the best performance. This is presented in \tableref{combined_results} and \figref{weights}.

\subsection{Counterfactual Explanations}
\begin{figure}[t]
    \centering
    \includegraphics[width=0.97\linewidth]{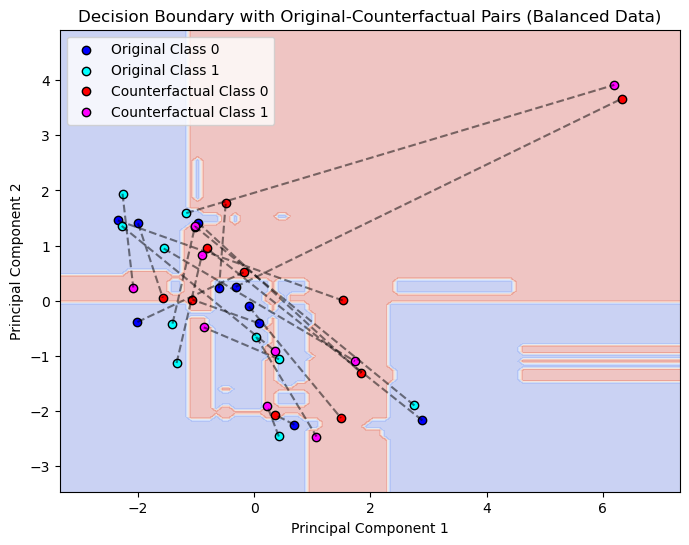}
    \caption{Decision boundary generated by random forest using 20 data points from the counterfactual examples. Here, Class 0 indicates an absence of MetS, and Class 1 indicates the presence of MetS.}
    \label{fig:cf_decision_boundary}
    \vspace{-2mm}
\end{figure}

The analysis of counterfactual pairs in PCA-reduced space reveals intricate patterns when using the Random Forest (RF) classifier. The visualization shows original instances clustered primarily in the central region (-2 to 2 on both principal components), while counterfactual instances exhibit wider dispersion, particularly extending into the positive range of Principal Component 1. The counterfactual transitions, represented by dashed lines, demonstrate varying lengths and directions of movement needed to cross the complex, nonlinear decision boundaries. The RF classifier creates multiple disjoint decision regions, visible as isolated ``island'' in the background coloring, highlighting its ability to capture local patterns in the data. This pattern is illustrated in \figref{cf_decision_boundary}, where both short and long transitions indicate varying degrees of feature space movement required to change classifications.

\subsection{Evaluation of the Counterfactual Explanations}
As presented in \tableref{overall_metrics}, the counterfactual analysis reveals distinct patterns in feature modifications required for outcome changes. The normalized average distance of 1.489 (±1.120) indicates that moderate to large perturbations were necessary to achieve counterfactual examples. On average, 2.054 (±1.070) features needed modification to change the prediction, representing 17.1\% of the total features. This suggests that the model relies on a relatively small but impactful set of key features for classification. The normalized average counterfactual distance was computed as:
\begin{equation}
    \vspace{-2mm}
    \mathbb{E} \left[ \frac{||x - x'||_0}{d} \right] = \frac{1}{N} \sum_{j=1}^{N} \frac{\sum_{i=1}^{d} \mathbf{1}(x_{i,j} \neq x'_{i,j})}{d}
\end{equation}
where $\mathbf{1}(\cdot)$ is an indicator function counting modified features.

\begin{table}[htbp]
    \centering
    \caption{Summary metrics of counterfactual examples. This table quantifies the feature modifications required to shift individuals from high-risk to low-risk MetS categories.}
    \begin{tabular}{ll}
        \toprule
        Metric & Value \\
        \midrule
        Average Normalized Distance & 1.489 \\
        Standard Deviation of Normalized  Distance & 1.120 \\
        Average Sparsity & 2.054 \\
        Standard Deviation of Sparsity & 1.070 \\
        Percentage of Features Changed & 17.1\% \\
        \bottomrule
    \end{tabular}
    \label{tab:overall_metrics}
    \vspace{-2mm}
\end{table}

\subsection{Feature-Specific Analysis}
As presented in \tableref{fsa}, the feature-specific analysis shows that metabolic markers were most frequently modified. Blood Glucose (50.3\%) and Triglycerides (46.7\%) exhibited the highest change rates, followed by Waist Circumference (42.9\%) and HDL (33.7\%). This pattern aligns with clinical understanding, as these features are crucial metabolic indicators. Notably, demographic features (Sex (0.1\%), Race (0.0\%), and socioeconomic factors (Income (1.7\%) showed minimal modifications, suggesting the model's predictions are robust to these characteristics. The relatively low modification rates for Age (8.9\%) and anthropometric measures like BMI (9.6\%) indicate that while these features are clinically relevant, they require less adjustment to affect the model's predictions. This analysis reveals that the model primarily relies on modifiable metabolic factors rather than fixed demographic characteristics for its predictions, which proves its usefulness in mitigating risk for metabolic syndrome through recommendations for achievable changes.

\begin{table}[htbp]
    \centering
    \caption{Feature-specific change rates in counterfactual analysis. This table highlights the most frequently modified features in counterfactual examples, with blood glucose and triglycerides being the most altered, reinforcing their critical role in MetS risk reduction.}
    \begin{tabular}{ll}
        \toprule
        Feature & Change Rate (\%) \\
        \midrule
        BloodGlucose & 50.3\% \\
        Triglycerides & 46.7\% \\
        WaistCirc & 42.9\% \\
        HDL & 33.7\% \\
        BMI & 9.6\% \\
        Age & 8.9\% \\
        UrAlbCr & 7.8\% \\
        UricAcid & 3.5\% \\
        Income & 1.7\% \\
        Sex & 0.1\% \\
        Albuminuria & 0.1\% \\
        Race & 0.0\% \\
        \bottomrule
    \end{tabular}
    \label{tab:fsa}
\end{table}

\section{Discussion}
This study tackles the challenge of predicting Metabolic Syndrome (MetS) amidst common healthcare dataset issues like class imbalance, data scarcity, and methodological inconsistencies, which can affect model reliability and clinical applicability.

Our research makes several important contributions to addressing these challenges. First, we developed a novel hybrid data balancing framework, MetaBoost, that chooses and combines a subset of SMOTE, ADASYN, and CTGAN and shows promise to perform better than other methods. This approach demonstrated robust results with an accuracy of 87.1\% and an F1 score of 0.868 when ADASYN and CTGAN are chosen, particularly noteworthy given the complexity of MetS prediction.

Our probabilistic analysis provides evidence for the clinical relevance of our approach. The high likelihood of elevated blood glucose (85.5\%) aligns with established clinical understanding of MetS, while the posterior probability analysis offers new insights into the relative importance of different risk factors. The counterfactual analysis further strengthens our findings by providing interpretable, actionable insights for clinicians. The fact that triglycerides and blood glucose emerged as the most frequently altered features in counterfactual examples (46.7\% and 50.3\% respectively) suggests these should be primary targets for intervention.

However, several limitations warrant discussion. First, while our hybrid data balancing approach showed promising results, its computational complexity might limit real-time applications. Second, the generalizability of our findings across diverse populations requires further validation. Third, the current implementation doesn't account for temporal variations in risk factors, which could be crucial for long-term risk assessment.

Future research should focus on validating these approaches in diverse clinical settings and exploring real-time deployment strategies. Additionally, incorporating temporal dynamics and investigating the impact of lifestyle factors could further enhance the model's predictive capabilities.

\section{Conclusion}
This study systematically evaluated and optimized machine learning models for predicting Metabolic Syndrome (MetS), addressing critical challenges such as class imbalance, data scarcity, and methodological inconsistencies. By leveraging advanced data balancing techniques, including random oversampling (ROS), SMOTE, ADASYN, and CTGAN, as well as a novel hybrid framework combining these methods, we demonstrated its potential to be used in other settings where it has the potential to perform well and possibly better than traditional approaches. The counterfactual analysis enhanced the interpretability of the solution, identifying triglycerides and blood glucose as the most frequently altered features in counterfactual examples.

The hybrid data balancing framework (MetaBoost), combining ADASYN and CTGAN, achieved an accuracy of 87.1\% and an F1 score of 0.868. This approach not only shows potential for model generalizability but also provides a unique, scalable solution for handling imbalanced datasets in healthcare analytics. Additionally, the counterfactual analysis can offer actionable insights for clinicians.

In conclusion, this study advances the methodological rigor of MetS prediction by integrating advanced ML techniques with probabilistic and counterfactual analyses. The findings underscore the potential of hybrid data balancing and interpretable ML models in addressing the global public health burden of metabolic syndrome.

\footnotesize{
\bibliographystyle{IEEEtran}
\bibliography{root.bib}
}

\end{document}